\pdfoutput=1
\documentclass[11pt]{article}
\usepackage[final]{acl}
\usepackage{times}
\usepackage{latexsym}
\usepackage[T1]{fontenc}
\usepackage[utf8]{inputenc}
\usepackage{microtype}
\usepackage{inconsolata}
\usepackage{graphicx}
\usepackage{subcaption}
\usepackage{multirow}
\usepackage{multicol}
\usepackage{booktabs}
\usepackage{array}
\usepackage{makecell}

\title{Improving Training Efficiency and Reducing Maintenance Costs via Language Specific Model Merging}

\author{
 \textbf{Alphaeus Dmonte\textsuperscript{1,2}},
 \textbf{Vidhi Gupta\textsuperscript{1}},
 \textbf{Daniel J Perry\textsuperscript{1}},
 \textbf{Mark Arehart\textsuperscript{1}}
\\
\\
 \textsuperscript{1}Qualtrics,
 \textsuperscript{2}George Mason University
\\
 \small{
   admonte@gmu.edu, \{vidhig, dperry, marehart\}@qualtrics.com 
 }
}

\begin{document}
\maketitle
\begin{abstract}
Fine-tuning a task-specific multilingual large language model (LLM) involves training the model on a multilingual dataset with examples in all the required languages. 
Updating one or more supported languages with additional data or adding support for a new language involves retraining the model, which can be computationally inefficient and creates a severe maintenance bottleneck. 
Recent research on merging multilingual multitask models has shown promise in terms of improved quality, but its computational and maintenance efficiency remains unstudied. In this work, we provide the first focused analysis of this merging strategy from an efficiency perspective, evaluating it across three independent tasks. 
We demonstrate significant efficiency gains while maintaining parity in terms of quality: this merging approach reduces the initial training time by up to 50\%. We also demonstrate that updating an individual language and re-merging as part of model maintenance reduces training costs by more than 60\%, compared to re-training the full multilingual model. We show this on both public and proprietary industry datasets confirming that the approach works well for industrial use cases in addition to academic settings already studied in previous work.
\end{abstract}

\section{Introduction}
Large Language Models (LLMs) are central to many NLP applications, and their performance is often enhanced by fine-tuning them on task-specific datasets. For multilingual applications, this typically involves training a single model on a combined, multilingual dataset. However, this ``retrain-all'' approach, while common, is computationally inefficient and creates a significant maintenance bottleneck. In a real-world enterprise setting, models are not static; they must be constantly updated with new data or expanded to support new languages and tasks. With the standard approach, adding or updating support for a single new language requires re-training the model on the entire combined dataset, an expensive and unscalable process.

\begin{figure}
    \includegraphics[width=\columnwidth]{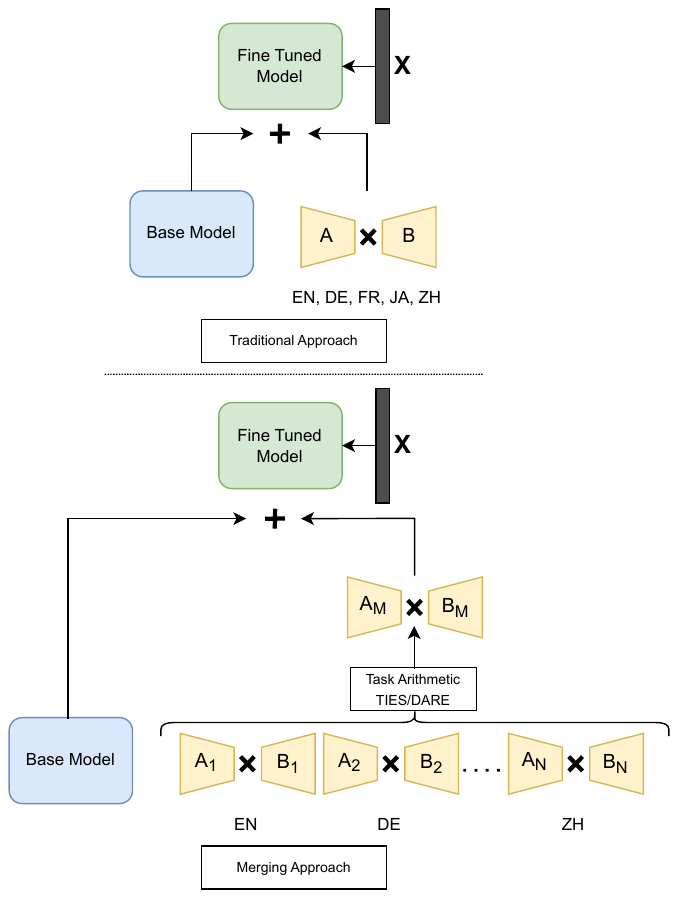}
    \caption{Traditional ``retrain-all'' training approach vs. Language Specific ``train-once, merge-as-needed'' approach.}
    \label{fig:merging}
\end{figure}

Model merging has recently emerged as a promising solution~\cite{parovic2024investigating, tao2024unlocking, zhao2025adamergex}. In this paradigm, models for individual tasks or languages are trained independently —a fast, concurrent process—and then merged into a single set of weights \cite{ortiz2023task, yadav2023ties, yu2024language}. While this technique has been explored in multi-task settings  and its multilingual performance has been touched upon \cite{zhao2025adamergex, pfeiffer2020mad, ahmadian2024mix}, its computational efficiency as a maintenance and training strategy is a critical, underexplored dimension.

This work provides an initial deep dive into language-specific model merging as a solution to the efficiency and maintenance problem in multilingual models. We directly compare the traditional ``retrain-all'' (combined dataset) approach against a ``train-once, merge-as-needed'' strategy. We move beyond just performance to quantify the significant savings in training time and cost.

\section{Related Work}

%Our work is positioned at the intersection of multilingual fine-tuning and model merging. Our contribution is focused around computational efficiency and model maintenance, while prior work focused on either zero-shot transfer or multi-objective performance.

\paragraph{Multilingual Fine-tuning and Model Merging:} 
The standard approach for creating task-specific multilingual models is to fine-tune a base model on a single, combined dataset containing all target languages~\cite{eisenschlos2019multifit, ladhak2020wikilingua, choenni2023languages, muennighoff2023crosslingual, indurthi2024improving}. 
While effective for initial model development, in an enterprise environment this becomes problematic because adding support for a new language or updating an existing one requires retraining the entire model. Our work directly addresses this inefficiency

With the increasing language model size, model merging has gained popularity to improve multitask model performance and model generalization~\cite{wortsman2022model, matena2022merging, stoicamodel}.

\paragraph{Cross-lingual Model Merging:} 
A subset of model merge research focuses on zero-shot cross-lingual transfer, which aims to enable a task in a target language without any labeled data for that language~\cite{zhao2025adamergex, pfeiffer2020mad, ahmadian2024mix}. These methods, while powerful, solve a different problem than ours.  Specifically,  \cite{ pfeiffer2020mad} proposes a modular framework that stacks two distinct adapter types: a language adapter (trained via MLM) and a task adapter. Zero-shot transfer is achieved by simply swapping the language-specific adapter at inference time.  In \cite{zhao2025adamergex}  also targets zero-shot transfer but uses adapter arithmetic. It calculates a ``language gap'' from a reference task and adds this gap to a source-language task adapter to create a new target-language adapter.
These approaches are designed for data-scarce scenarios. In contrast, our paper addresses the supervised scenario where training data is available for all languages, and the primary challenge is the computational cost of training and maintenance.

Our work is most closely related to \cite{ahmadian2024mix}, which also compares the ``combined training'' strategy against ``merge models'' strategy. However, their investigation has a fundamentally different goal: to find the optimal method for balancing two conflicting objectives (safety and general performance), limiting their analysis to quality and safety trade-offs. The primary motivation of our paper — computational efficiency, training cost, and model maintenance is not explored. Our work builds on their findings: we take the performance-parity of language-merging as a validated starting point and provide the first focused investigation into its economic and computational benefits. We demonstrate that it is not only a high-performing method but a practical and efficient solution for the real-world lifecycle of multilingual models in industry.

\section{Approach}

\subsection{Preliminaries}
\label{sec:preliminaries}
In this section, we give an overview of the merging techniques used in our experiments. Initial exploration of weight matrix concatenation and linear merging (task arithmetic) techniques produced inconsistent and irrelevant outputs. Hence, we further explore the following three widely used merging techniques.

\paragraph{TIES:} \citet{yadav2023ties} proposed Trim, Elect Sign, and Merge (TIES), a three step approach for merging models fine-tuned on multiple tasks. The top-k percent of each fine-tuned model's weights are retained, followed by sign selection, and finally merging the models by calculating the mean of the weights with the selected sign.

\paragraph{DARE:} Drop And REscale (DARE) \cite{yu2024language} first randomly sets certain weight values to 0, determined by a drop-rate $p$. The remaining weights are then scaled by a factor of $p/(1-p)$. The fine-tuned pruned models are then merged using an existing merging technique.

\paragraph{KnOTS:} \citet{stoicamodel} proposed Knowledge Orientation Through SVD (KnOTS), a precursor to model merging. The approach works by first concatenating the individual fine-tuned model weights layer by layer and then applying SVD over it to obtain a set of task-specific concatenated matrices. These matrices are then merged using an existing merging technique.

\begin{table*}[!ht]
\centering
\resizebox{\linewidth}{!}{
\begin{tabular}{lcccccc|cccccc|cccccc}
    \toprule
    \multirow{2}{*}{\textbf{Model}} 
      & \multicolumn{6}{c|}{\textbf{Summarization (BertScore)}} 
      & \multicolumn{6}{c|}{\textbf{Reasoning (Accuracy)}} 
      & \multicolumn{6}{c}{\textbf{Sentiment (F1)}} \\
    \cmidrule{2-19}
      & EN & DE & FR & JA & ZH & ALL
      & EN & DE & FR & JA & ZH & ALL
      & EN & DE & FR & JA & ZH & ALL \\
    \midrule
    
    $L8b_{COMB}$	 & \underline{0.839}	 & \textbf{0.834}	 & \textbf{0.837}	 & \underline{0.830}	 & 0.835	 & \underline{0.835}	 & 0.896	 & \textbf{0.840}	 & 0.754	 & 0.754	 & \underline{0.732}	 & \underline{0.795}	 & \underline{0.759}	 & \textbf{0.791}	 & 0.756	 & \underline{0.768}	 & \underline{0.675}	 & \textbf{0.750} \\
    $L8b_{INDV}$	 & 0.837	 & 0.817	 & 0.835	 & 0.814	 & \underline{0.836}	 & 0.828	 & 0.876	 & \underline{0.836}	 & \textbf{0.770}	 & 0.758	 & 0.724	 & 0.793	 & \textbf{0.791}	 & 0.755	 & 0.525	 & 0.775	 & 0.509	 & 0.671 \\
    $L8b\_DTS_{d=1.0}$	 & \textbf{0.840}	 & \underline{0.833}	 & \underline{0.836}	 & \textbf{0.836}	 & \textbf{0.838}	 & \textbf{0.836}	 & \underline{0.898}	 & 0.824	 & 0.756	 & 0.758	 & \textbf{0.736}	 & 0.794	 & 0.641	 & \underline{0.773}	 & 0.762	 & 0.758	 & \textbf{0.682}	 & 0.723 \\
    $L8b\_DT_{d=1.0}$	 & 0.811	 & 0.792	 & 0.797	 & 0.811	 & 0.813	 & 0.805	 & 0.874	 & 0.822	 & 0.752	 & \textbf{0.776}	 & 0.708	 & 0.786	 & 0.470	 & 0.747	 & \underline{0.769}	 & 0.459	 & 0.311	 & 0.551 \\
    $L8b\_TS_{d=1.0}$	 & \textbf{0.840}	 & \underline{0.833}	 & \underline{0.836}	 & \textbf{0.836}	 & \textbf{0.838}	 & \textbf{0.836}	 & \underline{0.898}	 & 0.824	 & 0.756	 & 0.758	 & \textbf{0.736}	 & 0.794	 & 0.641	 & \underline{0.773}	 & 0.762	 & 0.758	 & \textbf{0.682}	 & 0.723 \\
    $L8b\_T_{d=0.5}$	 & 0.836	 & 0.824	 & 0.828	 & \underline{0.830}	 & 0.832	 & 0.830	 & \textbf{0.908}	 & 0.832	 & \underline{0.760}	 & \underline{0.774}	 & 0.724	 & \textbf{0.800}	 & 0.651	 & 0.756	 & \textbf{0.774}	 & \textbf{0.778}	 & 0.659	 & \underline{0.724} \\
    \hline
\end{tabular}
}
\caption{Performance scores per task. COMB refers to the model trained on the combined dataset, INDV refers to the model trained and evaluated for each language independently and the other models refer to the merged models created by merging all 5 individual language adapters.  We show the baselines along with the best merged model for each merging technique (D=DARE, T=TIES, S=KnOTS). The ALL column refers to overall performance across all 5 languages.}
\label{tab:main-metrics}
\end{table*}

\subsection{Experimental Setup}

\subsubsection{Datasets}
To evaluate the effectiveness and generalization of multilingual model merging, we considered three independent tasks: Text Summarization, Commonsense Reasoning and Sentiment Analysis. We used the \textbf{WikiLingua}~\cite{ladhak2020wikilingua} dataset for summarization, \textbf{mCSQA}~\cite{sakai2024mcsqa} for reasoning and \textbf{MultilingualSentiment}~\cite{clapAI2024multilingualsentiment} for sentiment analysis. For each of the three tasks, we experimented with five languages: English (EN), German (DE), French (FR), Japanese (JA), and Chinese (ZH).

\subsubsection{Training Configurations}

We use Llama-3.1-8b-Instruct~\cite{grattafiori2024llama} as the base model for all tasks and languages. Each model was fine-tuned using LoRA~\cite{hulora} with r=64 and alpha=64 for 4 epochs. A learning rate of 2e-5 was used with a training batch size of 8 and maximum sequence length of 8196. For each task we use 500 validation and 500 test examples for evaluation. The Text Summarization task used 3000 examples for training and the other two used 5000 examples each.

\subsubsection{Baselines}
For each task and language, we use two baselines: a model fine-tuned with a combined task-specific dataset (COMB) and an individual model trained on a task-specific, language-specific dataset (INDV). The combined dataset for a task includes the examples from the five individual language datasets. These baselines are used to ensure performance parity with model-merging, while assessing their computational efficiency.

\subsubsection{Merging}
We experimented with several combinations of the three merging techniques described in Section~\ref{sec:preliminaries}. More specifically, we used the following combinations: TIES, TIES-KnOTS, DARE-TIES, and DARE-TIES-KnOTS. Since DARE and KnOTS are precursors to other merging techniques, they cannot be used as standalone merging techniques. Two hyperparameters are used: weight vector that determines the amount of weight to be given to each fine-tuned model and density which determines the percentage of weight' values to be retained. For these hyperparameters, we use two combinations: (weights=1, density=1) and (weights=1, density=0.5). This resulted in 8 merged models for each task.

\subsubsection{Metrics}
We compute the macro-average F1-score, Precision and Recall for the Sentiment task to account for class imbalance. The Reasoning task is evaluated using multi-class accuracy, and Summarization is evaluated using ROUGE-1, ROUGE-L, and Bert-Score.

\section{Results and Discussion}

\subsection{Model Performance}

For the Summarization task, we see that the overall performance of the merged models is comparable to both the baselines, as seen in Table~\ref{tab:main-metrics}. Among the merged models, Llama-8b merged with TIES-KnOTS(TS) and DARE-TIES-KnOTS(DTS) have the best performance. Across the languages, the merged models outperformed the baselines on English, Japanese, and Chinese with BertScore improving between 0.1 to 0.6\%, as seen in Table~\ref{tab:main-metrics}. While the BertScores of the merged models were on par with the baselines, we see that merging can improve the ROUGE scores for some of the languages, as indicated in Appendix~\ref{sec:additional-results}

For the Reasoning task, we see that the overall accuracy of the merged models is comparable to that of the baseline models. Similar to Summarization, the baselines showed slightly better accuracy for German and French, while for English, Japanese, and Chinese, the merged models marginally outperformed the baselines. The difference in accuracy between the baselines and the best merged models ranges from 0.4 to 2.2\% absolute difference. The TIES model with a density of 0.5 outperforms the multilingual baseline by 0.5\%.

Unlike the other two tasks, for Sentiment Analysis, the model trained on the combined dataset achieves the best performance, however, some of the merged models outperform individual language-specific models. The performance of the merged model for French, Japanese, and Chinese is slightly higher than that of the individual and multilingual models. However, there is an absolute difference of 1.7\% between the combined baseline and the best merged model for German, while for English, there is a significant performance difference, of the magnitude of 15\% between the best merged model and the best baseline. The overall lower performance of the merged models can be attributed to the lower performance of these models on English. 

From Table~\ref{tab:main-metrics} we see that the models merged with DARE-TIES(DT) and TIES(T) with density=1 have the same metric scores across all tasks. This is because, when the density is set at 1, none of the weights are pruned and since the post-pruning steps are the same between DARE and TIES, we observe similar performance for both these approaches.

Overall, merged-models seem to be on par with the combined model in terms of model performance. We even observe improvements in certain languages with the Summarization and Reasoning tasks. However, with Sentiment Analysis, merged model underperforms the combined model. This indicates that the effect of merging maybe task-specific. We hypothesize that the limited label space with classification tasks like sentiment may not work well with merging techniques, while for generative tasks like summarization and reasoning where the label set is varied, merging techniques may have more of a positive influence on the performance. We plan on testing this hypothesis in future work.

\begin{table}[!ht]
\centering
\resizebox{\columnwidth}{!}
{
\begin{tabular}{lcc}
    \toprule
    \multirow{3}{*}{\textbf{Model}} 
      & \multicolumn{2}{c}{\textbf{Training Time}} \\
    \cmidrule{2-3}
      & \makecell{Combined \\ Model} & \makecell{Merged \\ Model} \\
    \midrule

    Initial Setup & 3.4h & 2.2h (35.3\% $\downarrow$) \\
    Update/Add Language & 3.8h & 1.0h (73.7\% $\downarrow$) \\
    \bottomrule
\end{tabular}
}
\vspace{1em} 
\resizebox{\columnwidth}{!}
{
\begin{tabular}{lcc}
    \toprule
    \multirow{3}{*}{\textbf{Model}} 
      & \multicolumn{2}{c}{\textbf{Training Cost}} \\
    \cmidrule{2-3}
      & \makecell{Combined \\ Model} & \makecell{Merged \\ Model} \\
    \midrule

    Initial Setup & \$113.4 & \$107.1 (5.6\% $\downarrow$) \\
    Update/Add Language & \$119.7 & \$31.5 (73.7\% $\downarrow$) \\
    % \midrule
    % Total & \$233.1 & \$138.6 \\
    \bottomrule
\end{tabular}
}
\caption{Training Time and Cost Improvements.}
\label{tab:comp-eff}
\end{table}

\subsection{Computational Efficiency}

Table~\ref{tab:comp-eff} shows the efficiency gains when we use model-merging. During the initial setup when the model is being trained we observe that the training time reduces by 35\% when we use merging techniques as individual language models can be trained in parallel. We also see marginal reductions in training costs during this initial setup phase.

We also observe significant efficiency gains in the model maintenance phase. We conducted an ablation (Section~\ref{sec:language-update}) where we add additional EN examples to the training dataset. The combined model had to be re-trained entirely whereas with the merged model we only retrained the EN specific adapter and re-used the adapters of the other languages as is. Both the training time and training cost shows more than 70\% reductions when we used model-merging as compared to the traditional combined training approach. This suggests that, multilingual model merging achieves on-par performance to models trained on combined dataset with significant gains in computational efficiency and reduction in maintenance costs.

\section{Ablations}
\label{sec:ablations}
To further understand the advantages and disadvantages of language-specific model merging, we conduct additional ablations on the tasks. More specifically, we try to understand the impact of conducting language specific updates, language grouping, and the performance difference of merging smaller LLMs. We also merge task-specific language-specific models to evaluate how the performance is affected across the tasks and languages. This section discusses the language specific updates and impact of model sizes in detail and the rest of the ablations are available in Appendix~\ref{sec:additional-ablations}

\begin{table}[!ht]
\centering
\resizebox{\columnwidth}{!}
{
\begin{tabular}{lccccc}
    \toprule
    \multirow{2}{*}{\textbf{Model}} 
      & \multicolumn{5}{c}{\textbf{Sentiment (F1)}} \\
    \cmidrule{2-6}
      & EN & DE & FR & JA & ZH \\
    \midrule
    $MERGED_{Best}$ & 0.651	& 0.756	& 0.774	& 0.778 & 0.659 \\
    % $TIES_{EN-Updated}$ & 0.816 & 0.835 & 0.831 & 0.837 & 0.836 & - & - & - & - & - & 0.684 & 0.781 & 0.771 & 0.782 & 0.666 \\
    $TIES_{EN-Updated}$ & 0.684 & 0.781 & 0.771 & 0.782 & 0.666 \\
    \bottomrule
\end{tabular}
}
\caption{Results for the merged model with updated EN adapter.}
\label{tab:ablation-results-updated-en}
\end{table}

\subsection{Language Specific Update}
\label{sec:language-update}
To understand the impact of updating the adapter for a single language on the merged model, we retrain the adapter of a specific language using additional data. We use sentiment analysis as a case study for this experiment. For this task, since English had the lowest performance among all the languages, we retrain the English adapter with an additional 5,000 examples. Merging the updated English adapter with the adapters for the other four languages showed an improved F1 score on English compared to the best merged model. We further observe that updating the English adapter not only improved the performance of the merged model for English, but we also see a performance improvement in three other languages, as seen in Table~\ref{tab:ablation-results-updated-en}. The updated model was unable to surpass the baseline performance, however we observe improvement over the initially created merged model. All these results suggest that updating a single language adapter can improve the overall model performance at a fraction of the costs (Table~\ref{tab:comp-eff}).

% however despite the improvements we were not able to surpass the baseline performance for sentiment analysis.

\subsection{Impact of Model Size}
We investigate the impact of merging with smaller LLMs via the Llama-3.2-3b-Instruct~\cite{grattafiori2024llama} model. We focus on smaller LLMs as they are of interest in an enterprise setting to improve latency and cost. Similar to the Llama-8b experiments, we train language-specific models for sentiment analysis and summarization. For each task, we merge the language-specific models using the best hyperparameters and merging methods from the initial experiments. The Llama-3b model showed similar behavior as the Llama-8b model on merging, as seen in Table~\ref{tab:model-size-tab}. For summarization, the merged Llama-3b model achieved BertScore of 0.826 which was on par with the combined Llama-3b model's score of 0.831. For sentiment analysis, the F1 score of the Llama-3b merged model at 0.702 is slightly lower than the combined Llama-3b model's F1 score at 0.742. For both these tasks, this pattern is consistent with what we observed with the Llama-8b model where the summarization performance was on-par between the Combined and Merged model, whereas with Sentiment Analysis the Merged Model score was slightly lower than the Combined Model. The Llama-3b model is slightly worse when compared to the Llama-8b model, which is expected given that the Llama-3b model has significantly lower number of parameters. This experiment indicates that model merging is size agnostic and can be applied to LLMs of different sizes, however, the absolute performance may vary depending on how small or large the LLM is.

\begin{table}[!ht]
\centering
\resizebox{\columnwidth}{!}
{
\begin{tabular}{lcc}
    \toprule
    \multirow{4}{*}{\textbf{Base Model}} 
      & \multicolumn{2}{c}{\textbf{\makecell{Task: Summarization \\ (BertScore)}}} \\
    \cmidrule{2-3}
      & \makecell{Combined \\ Model} & \makecell{Merged \\ Model} \\
    \midrule
    Llama-3.1-8b-Instruct & 0.835 & 0.837 \\
    Llama-3.2-3b-Instruct & 0.831 & 0.826 \\
    \bottomrule
\end{tabular}
}
\vspace{1em} 
\resizebox{\columnwidth}{!}
{
\begin{tabular}{lcc}
    \toprule
    \multirow{4}{*}{\textbf{Base Model}} 
      & \multicolumn{2}{c}{\textbf{\makecell{Task: Sentiment \\ (F1)}}} \\
    \cmidrule{2-3}
      & \makecell{Combined \\ Model} & \makecell{Merged \\ Model} \\
    \midrule
    Llama-3.1-8b-Instruct & 0.750 & 0.724 \\
    Llama-3.2-3b-Instruct & 0.742 & 0.702 \\
    % \midrule
    % Total & \$233.1 & \$138.6 \\
    \bottomrule
\end{tabular}
}
\caption{Performance comparison across model sizes.}
\label{tab:model-size-tab}
\end{table}

% \begin{figure}[!ht]
% \centering
% \includegraphics[width=\columnwidth]{images/model_size_comp.pdf}
% \caption{Performance comparison across model sizes.}
% \label{fig:model-size-plot}
% \end{figure}

\section{Case Study}
To further understand the effectiveness of this technique for enterprises, we undertake a case study using a proprietary dataset. This task is similar to summarization, where an LLM processes unstructured data to identify relevant themes and provide supporting examples extracted from the input. It is supported in five languages, English, Spanish, German, French, and Japanese. The primary metric used is Aggregated Hallucination Rate\footnote{Lower hallucination rate is better.}, which computes the proportion of the number of LLM-generated examples that are not in the input. Similar to prior experiments, we first fine-tune language-specific models and a single multilingual model. We use the individual and the multilingual models as the baselines. Llama-3.1-8b-Instruct~\cite{grattafiori2024llama} is used as the base model.

\begin{figure}[!ht]
\centering
\includegraphics[width=\columnwidth]{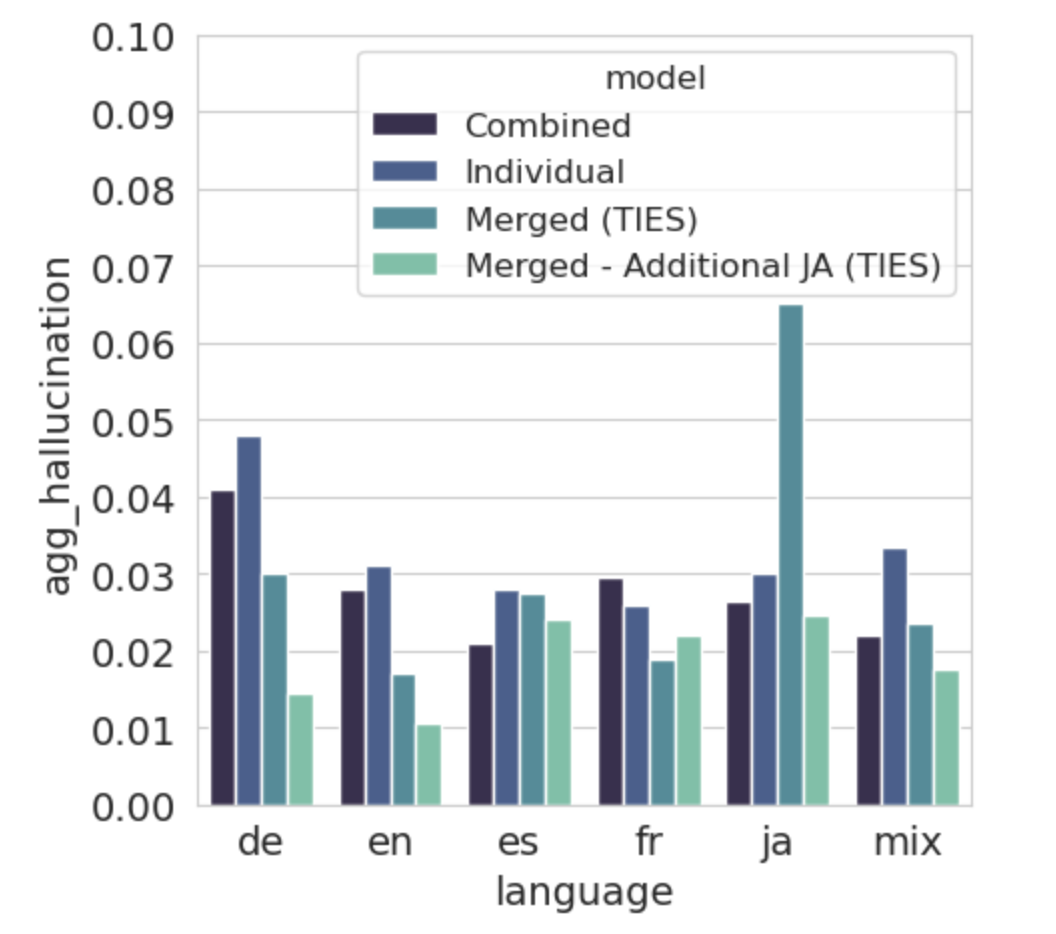}
\caption{The aggregated hallucination rate across the languages (lower is better). The plot shows the scores for four models, two baselines, and the best performing merged model TIES. The scores for the model merged with updated Japanese data are also reported. The 'mix' language refers to having more than 1 language in the input that needs to be summarized.}
\label{fig:hallucination-rate}
\end{figure}

\begin{table}[!ht]
\centering
\resizebox{\columnwidth}{!}
{
\begin{tabular}{lcc}
    \toprule
    \multirow{3}{*}{\textbf{Model}} 
      & \multicolumn{2}{c}{\textbf{Training Time}} \\
    \cmidrule{2-3}
      & \makecell{Combined \\ Model} & \makecell{Merged \\ Model} \\
    \midrule

    Initial Setup & 45h & 22.5h (50\% $\downarrow$) \\
    Update/Add Language & 54.5h & 20.5h (62.4\% $\downarrow$) \\
    \bottomrule
\end{tabular}
}
\vspace{1em} 
\resizebox{\columnwidth}{!}
{
\begin{tabular}{lcc}
    \toprule
    \multirow{3}{*}{\textbf{Model}} 
      & \multicolumn{2}{c}{\textbf{Training Cost}} \\
    \cmidrule{2-3}
      & \makecell{Combined \\ Model} & \makecell{Merged \\ Model} \\
    \midrule

    Initial Setup & \$1416 & \$1400 (1.1\% $\downarrow$) \\
    Update/Add Language & \$1717 & \$645 (62.4\% $\downarrow$) \\
    % \midrule
    % Total & \$3133 & \$2045 \\
    \bottomrule
\end{tabular}
}
\caption{Training Time and Cost Improvements Observed in the Case Study.}
\label{tab:case-study-tvc}
\end{table}

% \begin{figure}[h]
% \centering
% \includegraphics[width=\columnwidth]{images/case_study_tvc2.pdf}
% \caption{Training Speed and Cost Improvements}
% \label{fig:case-study-tvc}
% \end{figure}

We merge the language-specific models using the three techniques described in Section~\ref{sec:preliminaries}. For this experiment, we assign differing weights to each language model based on the relative importance of these languages for the business. As seen in Figure~\ref{fig:hallucination-rate}, experimental results showed that the merged models achieved a comparable performance or improved the performance over the baselines for all languages except Japanese. We observed that Japanese had the highest hallucination rate among all the languages. Hence we retrained the Japanese model with more training data. Merging the retrained Japanese model with other language adapters not only improved the performance of the merged model on Japanese, but we also observed improved performance for other languages like English and German. This supports our initial observation from Section~\ref{sec:language-update} that performance improvement may propagate across languages.

The experiment further demonstrates the effectiveness of language-specific model merging. As seen in Table~\ref{tab:case-study-tvc} model merging allows us to save on training time and eventual training costs without compromising on the model performance. We were able to update the Japanese adapter at 37.6\% of the cost via model merging as compared to the traditional approach. Apart from computational efficiency, merging allows the hyperparameters for each language to be tuned separately depending on the business needs, giving more language-specific control.

\section{Conclusion}
In this work we utilize existing language model merging techniques in a multilingual setting. Specifically, we use three techniques TIES, DARE, and KnOTS, and experiment on three public datasets. Results indicate that TIES merging overall had the best performance across the three tasks. The experiments demonstrate that the ``train-once, merge-as-needed'' approach achieves comparable performance to ``retrain-all'' approach, while offering significant savings in terms of training time and costs. Additional experiments on a proprietary datasets validates the findings and show similar training time and cost savings as the public datasets.

As a part of the future work, we plan to explore additional LLM sizes and families, and investigate ways to improve individual adapter weight selection. We further plan to perform hyperparameter tuning to improve the task specific performance, as well as leverage additional merging techniques.

\section*{Limitations}

This paper focuses on a single model family, other open source models may exhibit different performance characteristics which was not addressed in our work. We also limit the experiments in this paper to 5 medium-to-high resource languages. It would be interesting to assess the impact on model performance and computational efficiency when we have to work with significantly larger number of languages and/or low-resource languages.

% The main limitation of this work is that we only evaluate one model family. However, other open-source models may achieve varied performance. We further limit the number of languages used to five. It would be interesting to investigate if significantly increasing the number of languages would show similar observations. Additionally, this work considers medium-to-high resource languages when evaluating the performance and efficiency of model merging, it would be interesting to assess this with low-resource languages.

% \section*{Ethical Considerations}

% TBD - Optional section 

% \section*{Acknowledgments}

% TBD - Optional section, only needed in the final version not during review

\bibliography{custom}

% \newpage

\appendix

\section{Appendix}

\begin{table*}[!t]
\centering
\resizebox{\linewidth}{!}{
\begin{tabular}{l|cccc|c|ccc}
  \toprule
  \multirow{2}{*}{\textbf{Model}} & \multicolumn{4}{c|}{{\bf Summarization}} & \multicolumn{1}{c|}{{\bf Reasoning}} & \multicolumn{3}{c}{{\bf Sentiment}} \\
  \cmidrule{2-9}
  & \textbf{ROUGE-1} & \textbf{ROUGE-2} & \textbf{ROUGE-L} & \textbf{BertScore} & \textbf{Accuracy} & \textbf{Precision} & \textbf{Recall} & \textbf{F1} \\
  \midrule
  $L8b_{COMB}$ & 0.012 & 0.002 & 0.012 & 0.835 & 0.795 & 0.750 & 0.751 & 0.750 \\
  $L8b_{INDV}$ & 0.011 & 0.001 & 0.010 & 0.828 & 0.793 & 0.691 & 0.670 & 0.671 \\ 
  $L8b\_DTS_{d=1.0}$ & 0.012 & 0.001 & 0.012 & 0.836 & 0.794 & 0.731 & 0.727 & 0.723 \\
  $L8b\_DTS_{d=0.5}$ & 0.010 & 0.001 & 0.010 & 0.830 & 0.800 & 0.720 & 0.717 & 0.712 \\
  $L8b\_DT_{d=1.0}$ & 0.007 & 0.001 & 0.007 & 0.805 & 0.786 & 0.654 & 0.646 & 0.646 \\
  $L8b\_DT_{d=0.5}$ & 0.007 & 0.001 & 0.007 & 0.798 & 0.768 & 0.673 & 0.670 & 0.666 \\
  $L8b\_TS_{d=1.0}$ & 0.012 & 0.001 & 0.012 & 0.836 & 0.794 & 0.731 & 0.727 & 0.723 \\
  $L8b\_TS_{d=0.5}$ & 0.014 & 0.002 & 0.013 & 0.835 & 0.793 & 0.719 & 0.717 & 0.711 \\
  $L8b\_T_{d=1.0}$ & 0.007 & 0.001 & 0.007 & 0.805 & 0.786 & 0.654 & 0.646 & 0.646 \\
  $L8b\_T_{d=0.5}$ & 0.011 & 0.001 & 0.010 & 0.830 & 0.800 & 0.729 & 0.727 & 0.724 \\
  \bottomrule
\end{tabular}
}
\caption{Performance scores of all models across all languages for Summarization, Reasoning, and Sentiment Analysis.}
\label{tab:all-metrics}
\end{table*}

\subsection{Additional Results}
\label{sec:additional-results}

This section provides Table~\ref{tab:all-metrics} showing the values of all the computed metrics for each of the merging techniques and hyperparameter combinations we experimented with.

% and Figure~\ref{fig:lang-wise-plot} that showcases the language wise model performance of the best merged model per task against the combined and individual baselines, highlighting the on par performance observed for Summarization and Reasoning and the exceptions with Sentiment Analysis.

% \begin{figure*}[!ht]
% \centering
% \includegraphics[width=\linewidth,height=15cm]{images/lang_wise_model_perf.png}
% \caption{Language Wise Model Performance of the best merged model vs combined vs individually trained}
% \label{fig:lang-wise-plot}
% \end{figure*}

\begin{figure*}[!t]
\centering
    \begin{subfigure}{0.32\textwidth}
        \includegraphics[width=\linewidth]{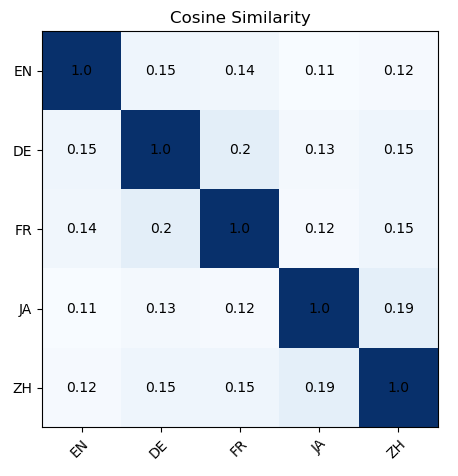}
    \caption{Summarization}
    \end{subfigure}
    \begin{subfigure}{0.32\textwidth}
        \includegraphics[width=\linewidth]{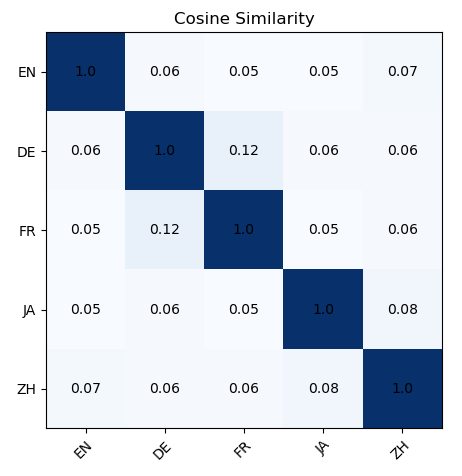}
    \caption{Reasoning}
    \end{subfigure}
    \begin{subfigure}{0.32\textwidth}
        \includegraphics[width=\linewidth]{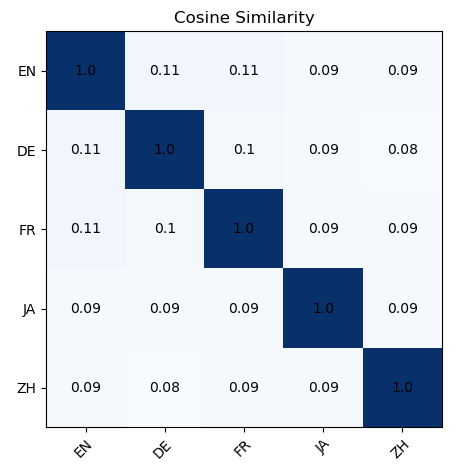}
    \caption{Sentiment Analysis}
    \end{subfigure}
    
    \caption{Cosine similarity between language vectors for Summarization, Reasoning, and Sentiment Classification tasks. The similarity score is computed between each language pair.}
    \label{fig:cosine-similarity}
\end{figure*}

\subsection{Language Vector Orthogonality}
\label{sec:language-vector}

\citet{ilharcoediting} show that the improved performance of the merged model on different tasks can be attributed to lower interference among the merged task vectors. To investigate if there is interference between the vectors for the different languages, we check the orthogonality among the language vectors for all three tasks. Language vector for a specific language is obtained by computing the difference between the weights of the fine-tuned model and the base model. In our case, since we use LoRA, that do not directly update the base model weights, we consider the product of the weight matrics A and B obtained after fine-tuning as the language vector for the specific language. The cosine similarity between any two language vectors is computed to check the orthogonality between them. The similarity matrices are shown in Figure~\ref{fig:cosine-similarity}.

We hypothesized that since all the languages are trained on the same task, the language vectors would be similar and hence they may not be orthogonal to each other. However, the cosine similarity computations reveal that the language vectors for any two languages have comparatively lower similarity, especially for related languages like English and German. This indicates that while all languages learn the same task, they may have different semantic learning spaces to adapt to a specific task. The amount of pre-training data used per language can also be a factor leading to the low similarity between languages. While the Llama-3 pretraining data contained a significant amount of English data, the data for other languages was minimal. Hence, to adapt to a specific task, during fine-tuning, the weight updates required for English compared to other languages are smaller. Other factors like language-specific semantics, syntactic structures, as well as model tokenization, can also influence the similarity between the vectors. For sentiment analysis and reasoning, the similarities are comparatively lower than those for summarization, indicating the task influence.

\subsection{Additional Ablations}
\label{sec:additional-ablations}

\begin{table*}[!ht]
\centering
\resizebox{\linewidth}{!}
{
\begin{tabular}{l|ccccc|ccccc|ccccc}
    \toprule
    \multirow{2}{*}{\textbf{Model}}
      & \multicolumn{5}{c|}{\textbf{Summarization}} 
      & \multicolumn{5}{c}{\textbf{Reasoning}} 
      & \multicolumn{5}{c|}{\textbf{Sentiment}} \\
    \cmidrule{2-16}
      & EN & DE & FR & JA & ZH
      & EN & DE & FR & JA & ZH
      & EN & DE & FR & JA & ZH \\
    \midrule
    $MERGED_{Best}$ & 0.840	 & 0.833	 & 0.836	 & 0.836	 & 0.838 & 0.908	 & 0.832	 & 0.760	 & 0.774 & 0.724	 & 0.651	 & 0.756	 & 0.774	 & 0.778 & 0.659 \\
    $TIES_{EN-Updated}$ & 0.816 & 0.835 & 0.831 & 0.837 & 0.836 & - & - & - & - & - & 0.684 & 0.781 & 0.771 & 0.782 & 0.666 \\
    $TIES_{EN-DE-FR}$ & 0.840 & 0.834 & 0.835 & - & - & 0.900 & 0.838 & 0.770 & - & - & 0.651 & 0.738 & 0.746 & - & - \\
    $TIES_{JA-ZH}$ & - & - & - & 0.836 & 0.838 & - & - & - & 0.740 & 0.732 & - & - & - & 0.771 & 0.667 \\
    \bottomrule
\end{tabular}
}
\caption{Results for the merged model with language cluster-based merging.}
\label{tab:ablation-results-cluster}
\end{table*}

\subsubsection{Language cluster-based merging}
To understand if merging the models based on shared language properties improves the task performance, we cluster the languages based on their shared vocabulary. Specifically we group them in two clusters: European languages, namely English, German, and French, and East Asian Languages, Japanese and Chinese. As seen in Table~\ref{tab:ablation-results-cluster}, for sentiment analysis, we observe a decrease in performance for German and French compared to the best merged model, while the performance was on par for the other three languages. For summarization we observed that the language cluster based merging achieved on-par performance to the best merged model. On the commonsense reasoning task, we see an increase in accuracy for German, French, and Chinese, while for there was a slight decrease in accuracy for English. For Japanese however, we saw that the accuracy decreased by 3.4\%. We can attribute this performance difference to the knowledge transfer during merging. When merging all languages, the merged models may inherit features from all the languages, while this transfer is limited with fewer languages. Moreover, the observations vary across tasks, indicating that merging the models based on language clusters may influence the performance differently based on the task. Overall, we did not observe a significant improvement with the language cluster-based model merging.

\subsubsection{Multitask-multilingual merging}

Previous works on model merging show that merging task-specific models overall improves the performance on all tasks. We therefore investigate how multilingual-multitask merging impact the performance across different tasks. To this end we merge language-specific and task-specific models. We consider two scenarios: merging all language-specific models across all tasks together, and first merging language-specific models for a task, followed by merging across tasks. In both these scenarios, the overall performance degrades for all the tasks, as shown in Table~\ref{tab:mt-mt-metrics}. The performance decrease is between 2-5\% for summarization and commonsense reasoning while for sentiment analysis, the F1 score drops by more than 5\%. Comparing the two scenarios, merging all the language models together performs best for summarization and reasoning, while for sentiment analysis, merging language-specific models followed by task-specific merging works best. While the results generally indicate lower performance, hyperparameter tuning for task-specific and language-specific merging may improve the overall performance, and we leave this for future exploration.

\begin{table}[!ht]
\centering
\resizebox{\columnwidth}{!}{
\begin{tabular}{lccc}
  \hline
  \bf Model & \bf Summarization & \bf Reasoning & \bf Sentiment \\
  & (BertScore) & (Accuracy) & (F1) \\
  \hline
  $BEST_{Merged}$ & 0.836 & 0.800 & 0.724 \\
  $TIES_{All}$ & 0.812 & 0.763 & 0.600 \\
  $TIES_{Each}$ & 0.810 & 0.755 & 0.665 \\
  \hline
\end{tabular}
}
\caption{Multitask Multilingual Merging, $TIES_{All}$ refers to merging all language and task adapters together; $TIES_{Each}$ refers to merging language adapters for each task creating a single task adapter followed by merging independent task adapters together.}
\label{tab:mt-mt-metrics}
\end{table}

\end{document}